\RequirePackage{fix-cm}
\documentclass[smallextended]{svjour3}       
\smartqed  
\usepackage{graphicx}
\usepackage{times}
\usepackage{fancyhdr,graphicx,amsmath,amssymb}
\usepackage[ruled,vlined]{algorithm2e}
\usepackage{mathtools}
\usepackage{cite}
\usepackage{amsmath,amssymb,amsfonts}
\usepackage{multirow}
\usepackage{textcomp}
\usepackage{xcolor}
\usepackage{float}
\usepackage{stackengine} 
\usepackage{tikz}

\newcommand\oast{\stackMath\mathbin{\stackinset{c}{0ex}{c}{0ex}{\ast}{\bigcirc}}}
\journalname{Neural Computing and Applications}
\begin{document}

\title{Signature Verification using Geometrical Features and Artificial Neural Network Classifier
}


\author{Anamika Jain         \and
       Satish Kumar Singh \and
       Krishna Pratap Singh 
}


\institute{Anamika Jain \at
              Indian Institute of Information Technology Allahabad\\
              \email{rsi2016005@iiita.ac.in}           
           \and
           Satish Kumar Singh \at
              \email{sk.singh@iiita.ac.in}
              \and
              Krishna Pratap Singh \at
              \email{kpsingh@iiita.ac.in}
}

\date{Received: date / Accepted: date}

\maketitle

\begin{abstract}
Signature verification has been one of the major researched areas in the field of computer vision. Many financial and legal organizations use signature verification as an access control and authentication. Signature images are not rich in texture; however, they have much vital geometrical information. Through this work, we have proposed a signature verification methodology that is simple yet effective. The technique presented in this paper harnesses the geometrical features of a signature image like center, isolated points, connected components, etc., and with the power of Artificial Neural Network (ANN) classifier, classifies the signature image based on their geometrical features. Publicly available dataset MCYT, BHSig260 (contains the image of two regional languages Bengali and Hindi) has been used in this paper to test the effectiveness of the proposed method. We have received a lower Equal Error Rate (EER) on MCYT 100 dataset and higher accuracy on the BHSig260 dataset.
\end{abstract}


%

\section{INTRODUCTION}

Biometric plays a vital role in the authentication of an individual in many financial institutions, and signatures are the most widely used modality for this purpose. 

Biometrics is used to identify or verify an individual digitally. The security applications that have been used in many financial and educational institutions using biometrics technology for decades\cite{8}. Biometrics are classified into two categories: physiological and behavioral\cite{14}. Physiological biometrics includes Face, Iris, Fingerprint, etc. and behavioral biometrics has signature, gait, etc..
 Authentication of the signatures are carried out manually and highly dependent on the mood of the verifier. Owing to its importance and unavailability of efficient offline verification methods, we have utilized signatures in this experiment\cite{1}\cite{review}. Depending upon the acquisition process, signature biometric is categorized into two modes, online and offline. In online mode, signatures are collected using tablets, electronic pads and have auxiliary informations like angle, pressure, pen up/down, etc. On the other hand, in offline mode the signatures are acquired on the sheet of paper with writing instruments. Later these sheets are digitized using the scanner and cropped to the signature content\cite{50yr}. These types of signature do not have any supportive information, and this makes the offline mode of the signature a challenging problem. 
 Signatures are easy to spoof with some practice, and this makes them vulnerable to forgery. There are two significant types of forgery reported in the literature, i.e., skilled and random forgery. In the skilled forgery, a person practices the genuine signature and tries to replicate it and in random forgery genuine signature of one user considered as the forged sample to the other signer.

\begin{figure}[!ht]
\includegraphics[width=\textwidth]{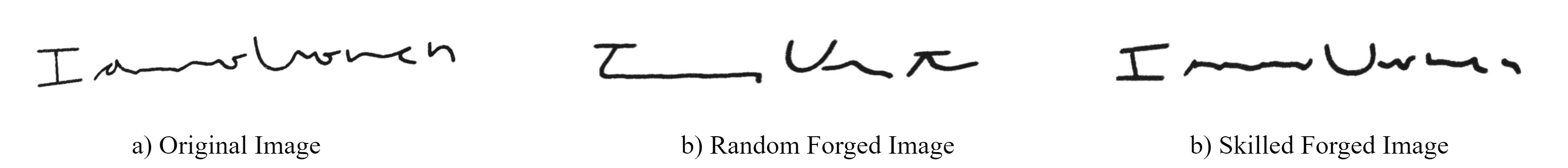}
        \caption{Sample Images}

\label{fig:sampleimages}
\end{figure}

Offline signature verification can be performed using two methods static and dynamic\cite{14}. Static approaches include geometric measurement, and dynamic approaches aim to estimate the information of the static image dynamically. 
Skillfully forged signatures have the same structure as the genuine signature, but the geometrical features of both the images are quite different. 



The rest of the paper is divided into five sections. Section II describes the state of the art method presented in the literature along with motivation of the proposed work. The proposed work and dataset used in this paper has described in section III. Section IV and V contain the experiment and results, respectively. In Section VI, conclusion and future work have been presented. 

\section{Related Work}
Signature verification is the method of access control in many organizations like financial organization for ages. This work presented here has focused on offline signature verification; hence, we discuss methods based on offline signatures only. 
There has been lot of work present in the literature\cite{1,2,3,4,6,7,8,9,10,11,12,13,14,19,121,22,amir,29,33,ali,36,108}. The signature verification methods has been divided in three categories based on the classifier used i.e. distance based\cite{5,36,108,amir}, SVM based\cite{4,29,33}, and Neural Network Based\cite{3,ali}. 

The authors have presented a clustering-based network for signature verification in \cite{5}. They utilized the regular and center moment-based features of the signature images. With the moment based features, authors have also calculated seven nonlinear features. The description of these seven features has presented in \cite{5}. With these features, clustering and combination of Kohonen Clustering Network (KCN) and fuzzy c-mean have applied. EFKCN has been used to update the learning rate to get the optimal learning rate. In \cite{36} authors presented a method based on the pixel intensity matching. Authors have used three different statistical classifiers, i.e., Decision tree\cite{37}, Naive Bayes\cite{38}, and euclidean distance. The classifier performed well when the limited training data was selected. Authors have used Discrete wavelet transform (DWT) to reduce the noise that can be introduced because of the acquisition process. The three classifiers have been trained using a ten-fold cross-validation manner. The writer independent process has the drawback of massive computation complexity. To overcome this limitation, the authors have proposed a method where they converted the multiclass problem into a bi-class (Genuine and Forgery) classification problem \cite{108}. They captured the non-similarity of the two samples depending on some threshold. They also made their method dataset independent. In this, if the model is trained on one dataset that can be utilized for other datasets also. They have used the power of contourlet transform(CT)\cite{109} to extract the discriminated features of the signature images. 
At the verification time system finds the dissimilarity measure between questioned and a reference image, and based on the decision threshold, the measure was selected, and the decision was taken. Authors have used a deep multimetric learning method in \cite{amir}. A distance metric has calculated for each class and trained with writer dependent and independent scenarios along with the SVM classifier. They have utilized the concept of transfer learning for feature extraction. The distances are used to identify the similarity and dissimilarity between the genuine and forged samples. Through this process, authors have generated more than one distance metric for each query sample. With the sum rule, these distances are fused to form one distance metric.


Despite many signature verification algorithms present in the literature, intraclass variation is still an issue for the researcher. To overcome this problem, authors \cite{4} proposed a method that is based on the selection of the best features that form a discriminate feature set. They have investigated global and local features in the signature image. In the local features, they have considered Centroid, slope, distance, and angle, and in a global feature, they utilized aspect ratio, area of signature in the image, height, width, and normalized length of the signature. Among all the features, the best features have been selected using the Genetic algorithm \cite{GA}. The resultant feature set has been given to the SVM for verification purpose. In \cite{29}, authors have proposed a method that was based on hybrid HOG features. Global statistics and key points have been extracted using the HOG features from the input signature image. A code-book was used to store these features and for the neighbors of the features. A coding was generated and stored back in the code-book. BRISK features were extracted from the signature image. These features detect the corner point in the signature image, and for each location, the Delaunay triangulation(DT)\cite{30} was applied. SVM was used to train on the features extracted using this method. In \cite{33}, authors have proposed an algorithm that was helpful in forensic cases. They showed the relation between the projection of feature space and training samples. They stated that the feature space is highly dependent on the number of training samples. The authors have used the unsupervised Archetypal analysis method for data analysis. This method works on a small number of samples, and the resulted archetypal was used to generate the feature space. The signature image and archetypes were used for the verification of the signature. Despite giving the excellent performance, this method has been dependent on the patch size. A smaller patch size offers a higher error rate.

In \cite{ali} authors has used artificial Neural Network for classification of the signature. They have extracted and used five global features, i.e., area, skewness, eccentricity, centroid, kurtosis. These features have been passed through the ANN for training. Based on the target results, the mean square error (MSE) will be adjusted. Their objective is to minimize the MSE as much as possible. Based on only the five features, authors have identified whether the query image is genuine or forged image. In \cite{115}, authors have presented a hybrid method that takes benefit from both types of settings, i.e., writer independent (WI) and writer dependent (WD). Since to train the writer dependent process large amount of data is required. At the starting phase of the training, when there is less number of samples, the writer independent process has been utilized. After the process starts and when there was a sufficient number of samples, the writer dependent process took over the independent process. The features that were extracted using the writer independent process worked as population representation. Directional probability density function and extended shadow code have been obtained using WI learning. While WI learning taking place, features using both genuine and forged samples have extracted, and these features were projected to the dissimilarity space using dichotomy transformation(DT)\cite{dt1}. With the boosting features selection process\cite{boost}, the feature was selected, and Ada-boost algorithm\cite{adaboost} was applied to learn the optimal decision boundary.


 
With the intuition that the signature images have distinguished geometrical features, we propose the signature verification method that uses geometrical properties from the signature image.  
 
From the literature, we have observed that Neural network based and SVM methods gives better performance as compared to the distance based methods. But SVM depending upon the size of the dataset can take more time in training the classifier, on the other hand neural network take less time in training. 



The proposed method is simple and effective that can be used for the purpose of the signature verification in real time applications.
 
\section{Proposed Methodology}
\begin{figure}[!ht]
    \centering
    \includegraphics[width=\linewidth]{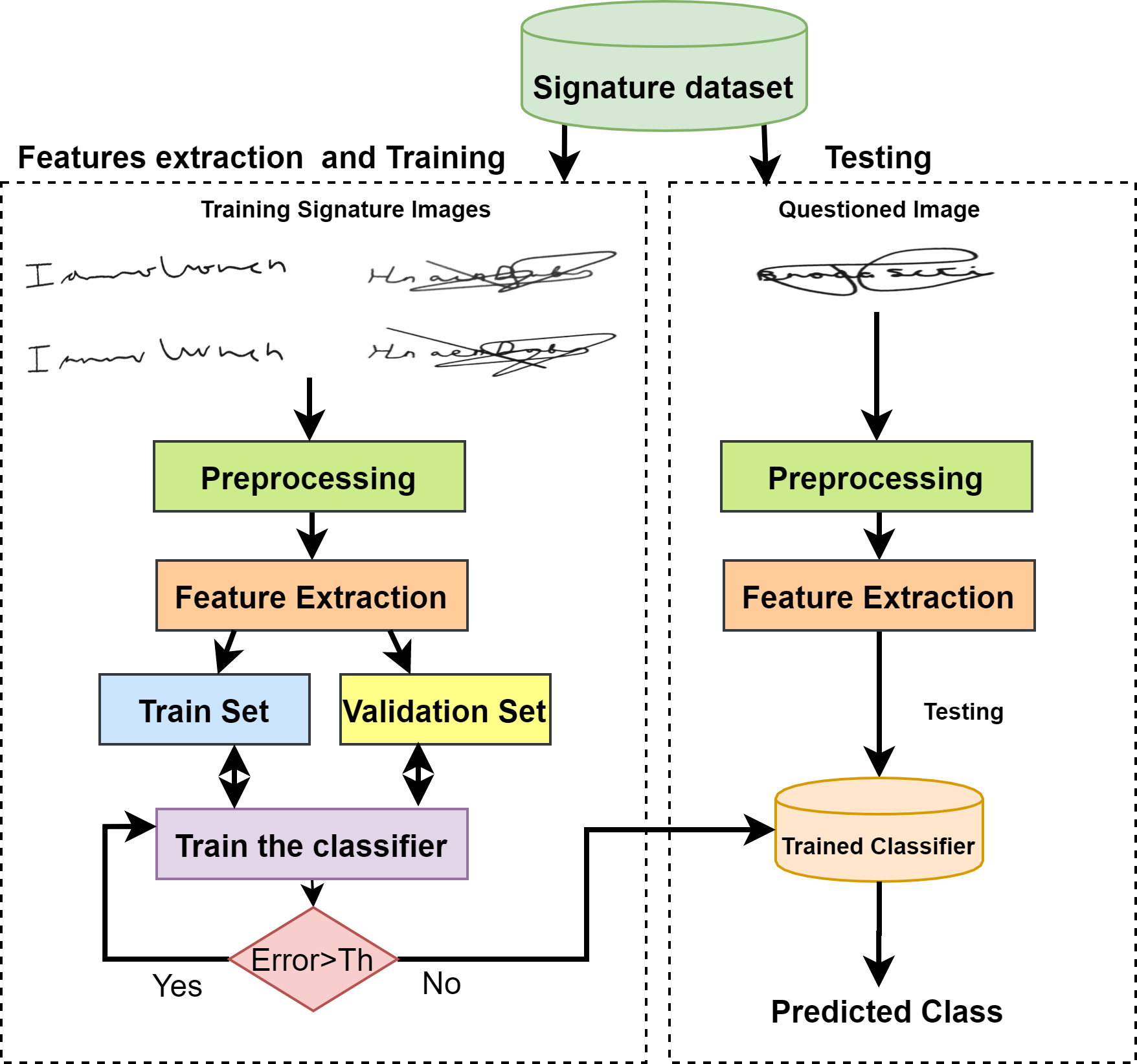}
    \caption{Brief Block diagram of the Proposed method}
    \label{fig:flowchartWhole}
\end{figure}

Figure \ref{fig:flowchartWhole} shows the brief block diagram of the proposed method. The proposed method has been divided into four parts i) Pre-processing, ii) Feature Extraction, iii) Training of the classifier, iv) Verification.

\subsection{Pre-processing}\label{preprocess}

Signature images have significant intra-class variations. To reduce the intra-class difference to some extent pre-processing is required before the extraction of the features for better feature extraction. These intra-class variations can be dependent on many intrinsic (mood, behavior, etc.) and extrinsic (ink type, ink color, acquisition process) parameters. In this experiment, we have used different datasets that are described in section \ref{dataset}. These datasets have different types of images. MCYT-100 is an online signature database, but for this experiment, we have utilized only signature images, not the auxiliary information. These images are not clear in terms of trajectory. To extract useful features, we have performed dilation operation with a structuring element of shape disk with radius 1. Dilation operation enhances the trajectories of the signature images in the MCYT-100 dataset. Dilation operation is defined as:

\begin{equation}
S \oplus E=\{x|(\hat E)_x \cap S \neq \phi\}
\label{dilation}
\end{equation}
Where S is the signature image and E is the structuring element, $\oplus$ is the dilation operation and $\hat E$ is the reflected structuring element.

Median filtering, binarization, and thickening has been applied on all the remaining datasets. Since MCYT-100 is a online dataset, it is not having any salt and paper noise because of the acquisition process, and it has already been binarized in the process of the dilation, so Median filtering and binarization process will not be applied on the MCYT-100 dataset. 
  
Due to the acquisition process, sometimes noise can be captured along with the signature images. To remove the noise, signature images have been passed through the median filter of the filter mask $3 X 3$, to eliminate the salt and pepper type of noise. The process of the filtering has been explained in the algorithm \ref{median}.






\begin{algorithm}[!ht]
\SetAlgoLined
\KwOut{Filtered Image $S_m$}
\KwIn{Signature Image (S), filter mask (M)}
$\begin{bmatrix}
S
\end{bmatrix}_{mXn}$ = $\begin{bmatrix}
\begin{array}{lll}
   \begin{bmatrix}
s
\end{bmatrix}_{11} & \cdots &  \begin{bmatrix}
s
\end{bmatrix}_{1n} \\
   \vdots & \ddots & \vdots \\
    \begin{bmatrix}
s
\end{bmatrix}_{m1} & \cdots &  \begin{bmatrix}
s
\end{bmatrix}_{mn}
\end{array}
\end{bmatrix}$

$\begin{bmatrix}
M
\end{bmatrix}_{3X3}$ = $\begin{bmatrix}
\begin{array}{lll}
  1 & 1 &  1 \\
  1 & 1 &  1 \\
  1 & 1 &  1 \\
\end{array}
\end{bmatrix}$

$\begin{bmatrix}
S_b^{i,j}
\end{bmatrix}_{3X3}$ = $\begin{bmatrix}
\begin{array}{lll}
  s_{i-1,j-1} & s_{i-1,j} &  s_{i-1,j+1} \\
  s_{i,j-1} & s_{i,j} &  s_{i,j+1} \\
  s_{i+1,j-1} & s_{i+1,j} &  s_{i+1,j+1} \\
\end{array}
\end{bmatrix}$

\For{i=1:m}{\For{j=1:n}{
$\begin{bmatrix}
f_b
\end{bmatrix}_{3X3}$ = $\begin{bmatrix}
S_b^{i,j}
\end{bmatrix}$.$\begin{bmatrix}
M
\end{bmatrix}$\\
$\begin{bmatrix}
\tilde{f_b}
\end{bmatrix}_{1X9}$= flatten$\begin{bmatrix}
f_b
\end{bmatrix}_{3X3}$\\
$\begin{bmatrix}
\tilde{f_b}
\end{bmatrix}_{1X9}$ = sort$\begin{bmatrix}
\tilde{f_b}
\end{bmatrix}$\\
$\begin{bmatrix}
f_b
\end{bmatrix}_{1X1}$ = median$\begin{bmatrix}
\tilde{f_b}
\end{bmatrix}$\\
$\begin{bmatrix}
S_m
\end{bmatrix}_{i,j}$ = $\begin{bmatrix}
f_b
\end{bmatrix}$
}}
\caption{Algorithm for median filtering}
\label{median}
\end{algorithm}

Where $S_m$ is the median filtered image, M is the $3X3$ filter mask. (m, n) is the size of the signature image.

The filtered image is segmented using Otsu's method. Otsu's method segment the foreground from the background using the thresholding method. The global threshold has been selected, such that the intra-class variance of the black and white pixels is minimized. The histogram has been calculated for the purpose of segmentation and selection of the threshold. 

\begin{equation}
\centering
\begin{split}
    p(h)=histogram(S_m)\\
    Th=SelectThreshold(p(h))
\end{split}
\end{equation}

\begin{equation}
    \begin{bmatrix}
    \boldsymbol{S_B}
    \end{bmatrix} = \left\{\begin{matrix}
0;  \quad  S_m(x,y)<Th\\ 
1;    \quad S_m(x,y)\geqslant Th
\end{matrix}\right.
\end{equation}

Where p(h) is the histogram of the median filtered image and Th is the threshold. Based on the probability densities of the foreground and background, the optimal threshold has been selected. According to the threshold, the image is converted into the binary image. 



\begin{figure}[!ht]
     \centering
\includegraphics[width=\linewidth]{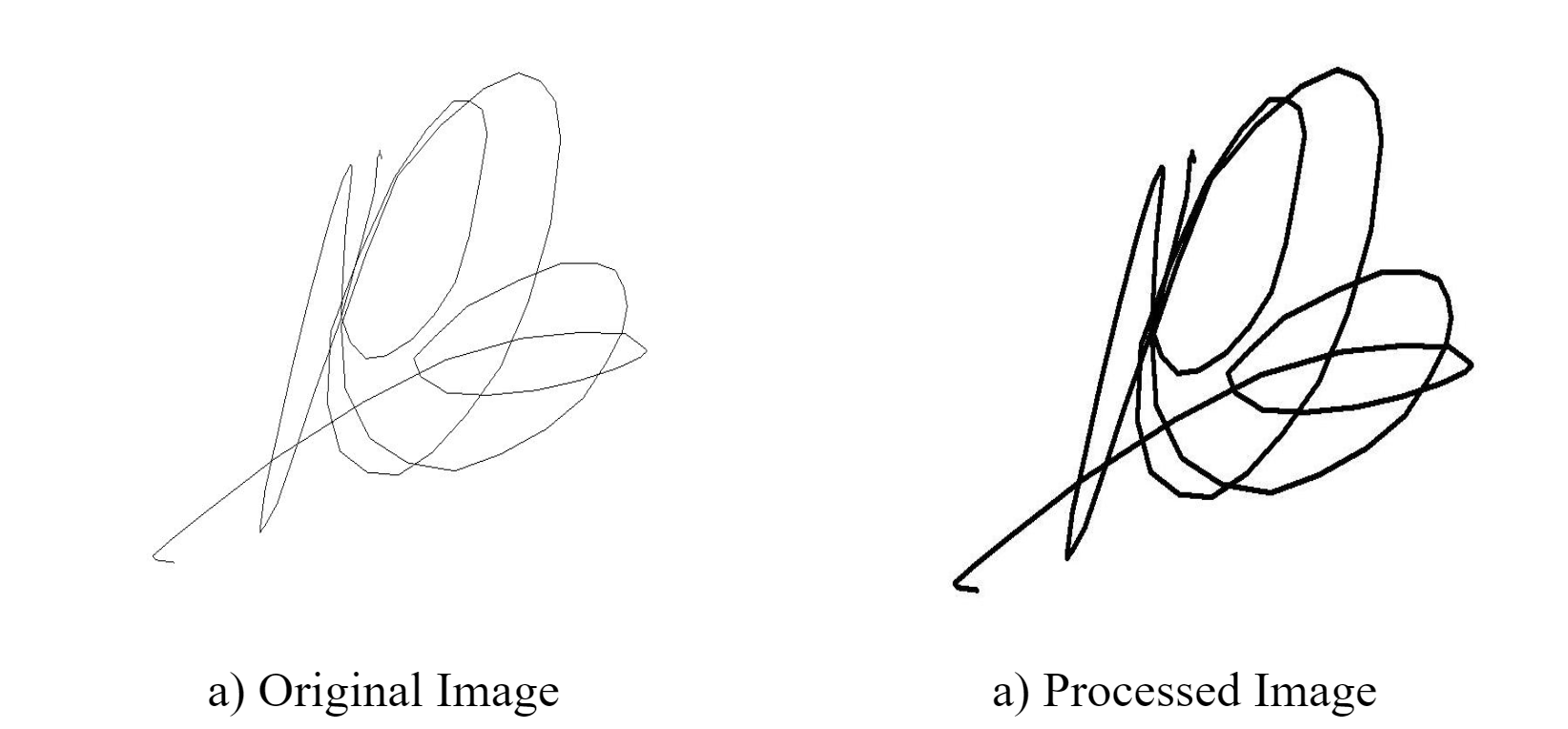}
\caption{Results of Pre-processing on MCYT-100 dataset}
\label{fig:mcytpreprocess}
\end{figure}

\begin{figure}[!ht]
 \includegraphics[width=\linewidth]{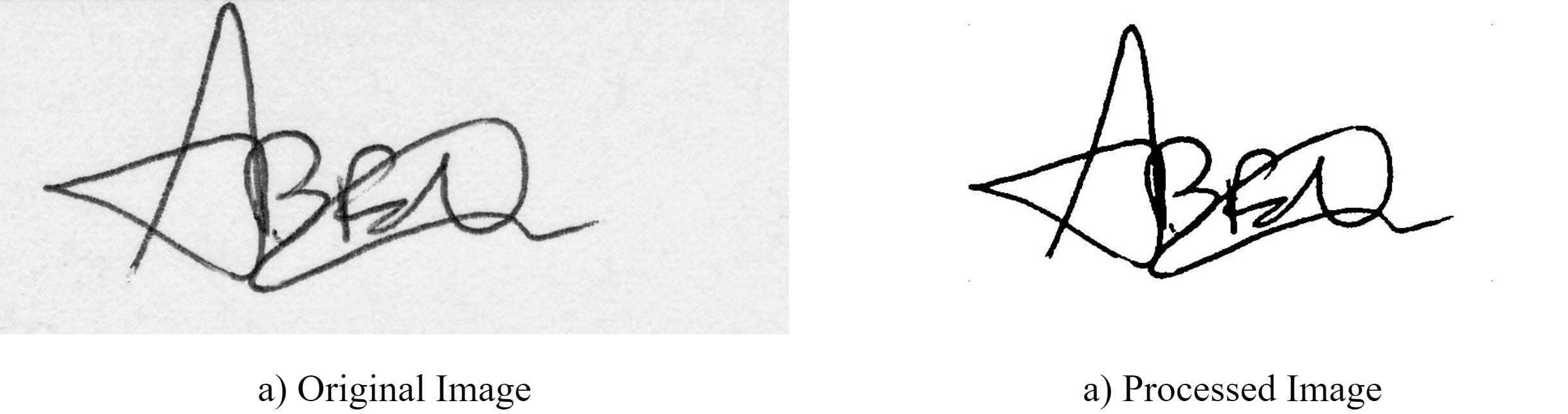}
\caption{Results of Pre-processing on MCYT-75 dataset}
\label{fig:mcyt75}
\end{figure}

Figure \ref{fig:mcytpreprocess} and Figure \ref{fig:mcyt75} shows the results of the processed MCYT dataset signatures.


\begin{figure}[!ht]
\includegraphics[width=\linewidth]{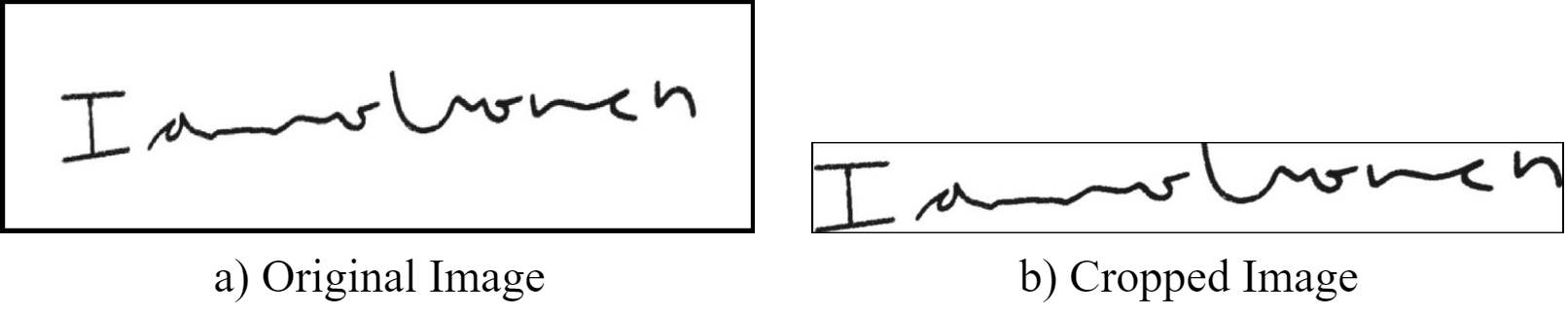}
\caption{Results of pre-processing}
\label{fig:fig}
\end{figure}

Binarized images form each dataset are cropped to the content based on the active pixel present in x and y-direction, as shown in figure \ref{fig:fig}. 

All the datasets are having different sizes of images. To make the similarity among the datasets, we have resized the images to a common dimension of size $192X256$ by bi-cubic interpolation while maintaining the aspect ratio to 3:4. 



\begin{equation}
\begin{bmatrix}
\boldsymbol{S_{mR}}
\end{bmatrix}=Bi\begin{bmatrix}
\boldsymbol{S_B}
\end{bmatrix}
\end{equation}
$S_{mR}$ is the pre-processed resized image and Bi is the bi-cubic interpolation.
After resizing, we have performed a morphological operation thickening to the infinite level. Thicken operation thickens the foreground by inserting pixels to the objects to form 8-connectivity. Thickening can be defined as 

\begin{equation}
S_{mR} \odot E=S_{mR} \cup (S_{mR} \oast E)
    \label{thicken}
\end{equation}

$S_{mR}\oast E = (S_{mR} \ominus E_1)-(S_{mR} \oplus (\hat E_2))$ where $\oast$ is the hit or miss transform and B is the structuring element and $E=(E_1,E_2)$ where $E_1$ belongs to the object and $E_2$ belongs to the background and S is the signature image. $\ominus$ and $\oplus$ are the erosion and dilation operations respectively.

These processed images are forwarded to the feature extraction process.


\subsection{Feature Extraction}
In this phase, two global and eight local features have been extracted from each signature image. The global feature set has been obtained on the whole signature image and includes the number of connected components and the number of active pixels in the signature image.

\begin{figure}[!ht]
    \centering
    \includegraphics[width=4cm,height=4cm]{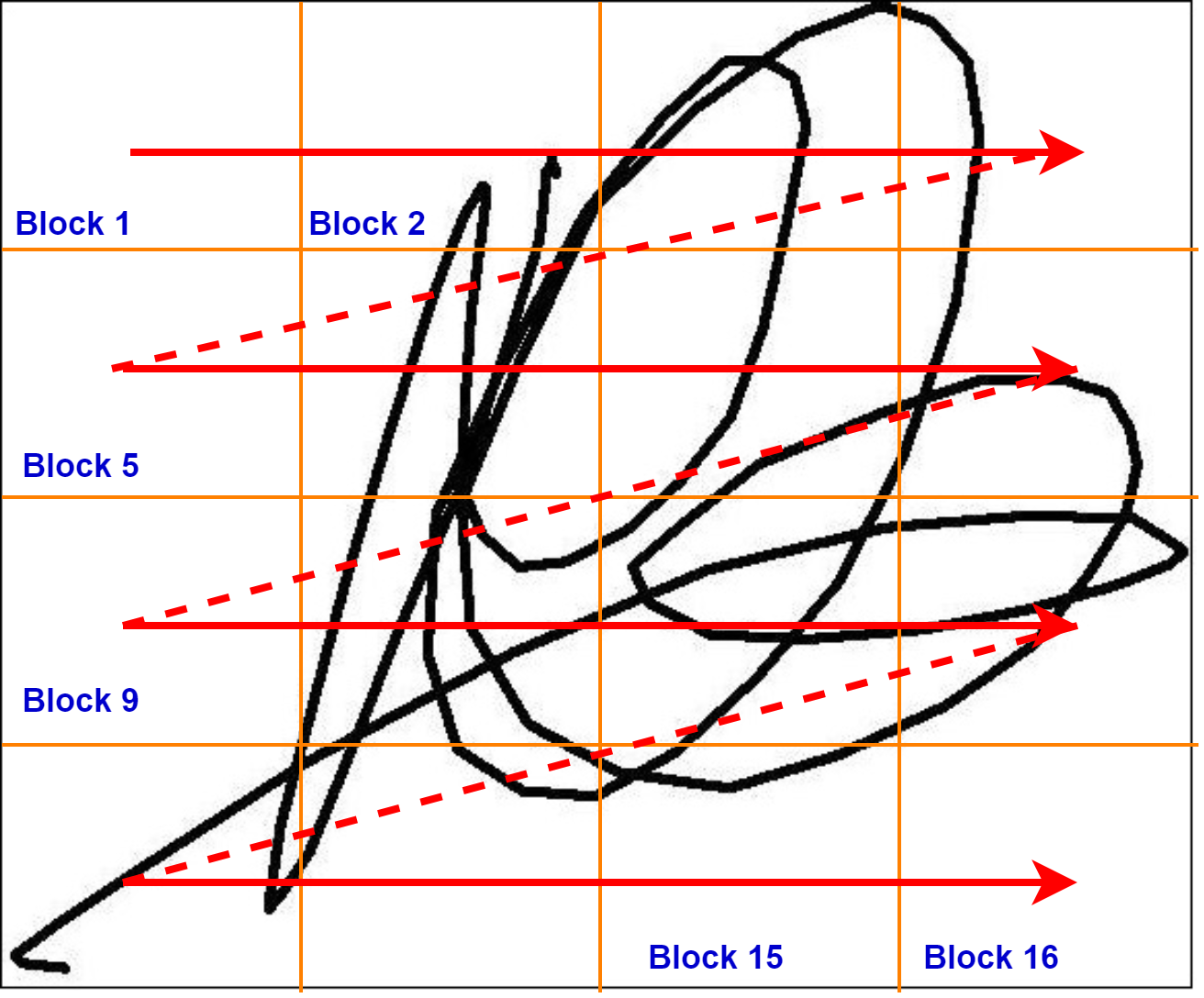}
    \caption{Division of Signature image in blocks}
    \label{fig:grid}
\end{figure}

Local features extracted from the blocks that have been obtained through the algorithm \ref{algo_block}. Signature image has been divided into the sixteen number of blocks, and the size of each block is (48x64), the resulted image has been shown in \ref{fig:grid}. Local features include co-ordinates of effective mass, the distance between the center of mass and effective mass, number of active pixels, number of connected components, isolated points, average height, and width of the signature in the signature image. The detailed description of these features has been presented in section \ref{lfeatures}.



                    
                                    

        

\begin{algorithm}[!ht]
\SetAlgoLined
\KwOut{Blocks from image($B_Q$)}
\KwIn{Signature Image(SigImage)}
X=0, Y=0\;Q=1, s=48\;t=64\;
 BlockDiv(SigImage,Q,s,t,X,Y)\;
 \{\\ 
    l=1;k=1\;
     \For{($i=X:BlockSize_x+X $)}
        {
            \For{($ j=Y:BlockSize_y+Y$)}
               {
                    \quad$Block_Q(l,k)$ = $SigImage_{(i,j)}$\;
                }                    
        }
        Y=Y+j\;
        \If {$Q\mod 4$==0}
            {
                X=X+s\;
                Y=0\;
            }
    \}       
    BlockDiv(SigImage,Q+1,s,t,X,Y)\;
 \caption{Algorithm to find Blocks from image}
\label{algo_block}
\end{algorithm}

\begin{figure*}[!ht]
    \centering
    \includegraphics[width=\linewidth]{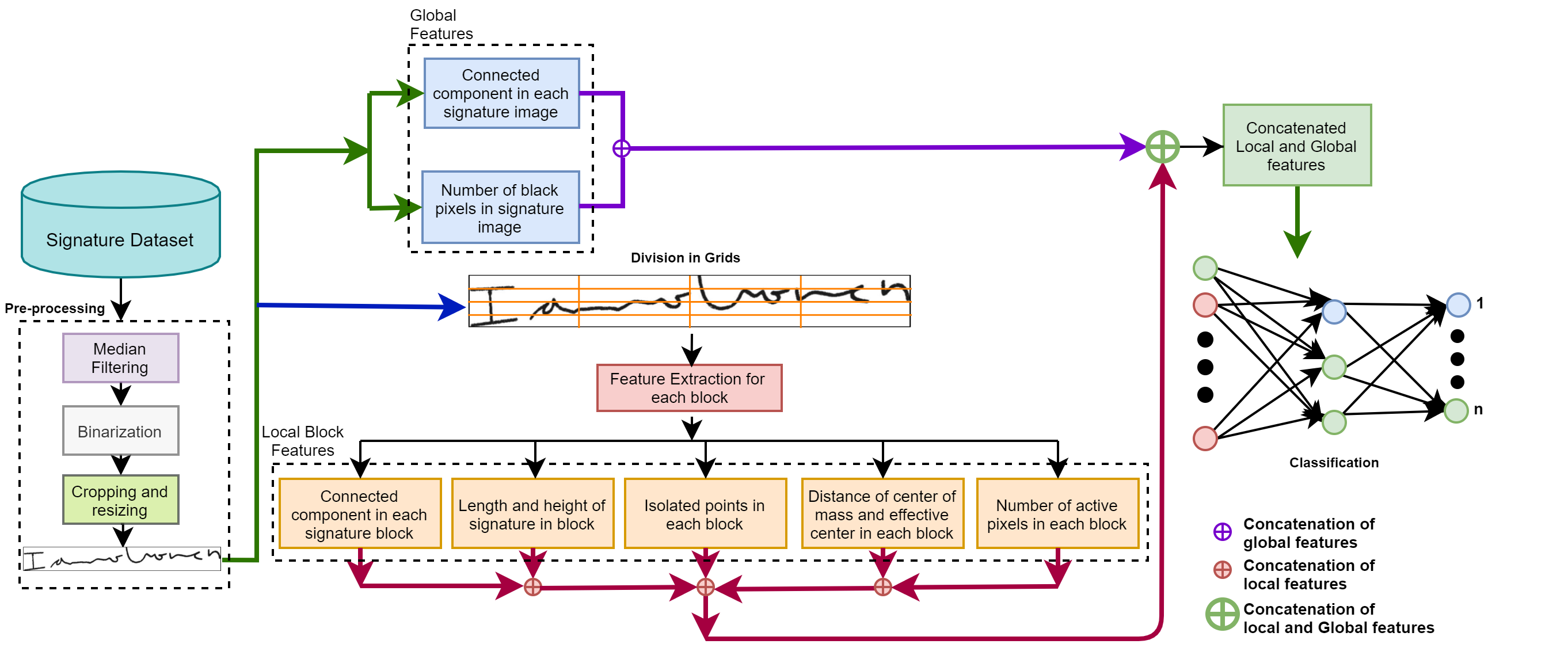}
    \caption{Flow Chart of proposed method}
    \label{fig:Proposed Method}
\end{figure*}

After the processing of all the blocks, the local features are concatenated to form a local feature set. 
For each signature image, local as well as global features(refer section \ref{gfeatures}) have been calculated. The local and global feature sets are concatenated to create a final feature vector. These features are fed to the artificial neural network for classification. The detailed flowchart and algorithm of the proposed method are shown in figure \ref{fig:Proposed Method} and algorithm \ref{algo}, respectively. Matrix decomposition of the proposed method is described in equation \ref{matrixForm}.

\begin{algorithm}[!ht]
\SetAlgoLined
\KwOut{Feature set}
\KwIn{Signature Image}
Find $X_{min},Y_{min}$ and $X_{max},Y_{max}$ to identify the region of interest\;
 Initialize rows of the blocks\;
 Initialize columns of the blocks\;
 \While{for each block}{
 Each the block is thickened to the infinite level\;
  find number of black pixels in the block($count_{pixels}^{block}$)\;
  \eIf{$count_{pixels}^{block}>20$}{1.Find center of mass($C_X^{mass}, C_Y^{mass}$) and effective center($C_X^{effec},C_Y^{effec}$)\;
   2. Find the distance between effective center of ($C_X^{effec},C_Y^{effec}$) and center of mass($C_X^{mass}, C_Y^{mass}$) of the block\;
      3. Find number of connected components\;
      4. Find number of isolated points\;
      5. Find length of the Signature in the block\;
      6. Find Average height of the signature in block\;
      7. Find number of pixels in the block\;
  }{
     Put zero at the place of that block\;
   }{
  $Feature Set_{local}$ = combine features of all the blocks\;
  }

 Thickened the original image to the infinite level\;
 $count_{black}$ = Calculate number of black pixel in the original image;\quad Feature of the whole image\;
 \vspace{0.1cm}
 $connected \quad Components$=Total number of connected components in the original signature image;  \t features of the whole image\;
 \vspace{0.1cm}
 Feature set = concatenate[$Feature Set_{local}$ , $count_{black}$, connected Components ]\;

 }
 
 \caption{Algorithm of the proposed method}
\label{algo}
\end{algorithm}

\begin{equation}
\begin{aligned}
\begin{bmatrix}
\textbf{S}
\end{bmatrix}= \begin{bmatrix}
{S}_{ij}
\end{bmatrix} =  \begin{bmatrix}
\begin{array}{lll}
   \begin{bmatrix}
s
\end{bmatrix}_{11} & \cdots &  \begin{bmatrix}
s
\end{bmatrix}_{1n} \\
   \vdots & \ddots & \vdots \\
    \begin{bmatrix}
s
\end{bmatrix}_{n1} & \cdots &  \begin{bmatrix}
s
\end{bmatrix}_{nn}
\end{array}
\end{bmatrix}\\ =
\begin{bmatrix} 
\begin{array}{ccc}
   \begin{bmatrix}
f_{11}
\end{bmatrix}_{1X8} & \cdots &  \begin{bmatrix}
f_{1n}
\end{bmatrix}_{1X8} \\
   \vdots & \ddots & \vdots \\
    \begin{bmatrix}
f_{n1}
\end{bmatrix}_{1X8} & \cdots &  \begin{bmatrix}
f_{nn}
\end{bmatrix}_{1X8}
\end{array}
\end{bmatrix}\\
=\begin{bmatrix}
\textbf{$F_1$}
\end{bmatrix}=
\begin{bmatrix}
F_1 \end{bmatrix} \begin{bmatrix}
 f_g
\end{bmatrix}
=\begin{bmatrix}
\textbf{F}
\end{bmatrix}_{1XN}
\end{aligned}
\label{matrixForm}
\end{equation}

In equation \ref{matrixForm}, \textbf{S} is the signature images and $S_{ij}$ are the elements of the signature image. [s] is one block of the image, and the size of the block is 48x64, f is the feature set of [s], where each block has the features of size 1x8. $F_1$ is the concatenated feature of all the blocks. $f_g$ is the global features extracted from the signature image, and \textbf{F} is the concatenated features of all the blocks and global features. Here N is the total size of the feature set after concatenation of the features from each grid.





The local and global features used in the proposed method are discussed below.
\subsubsection{Global Features}\label{gfeatures}
\begin{enumerate}
    \item \textbf{Number of Active pixel}: Each person has its style of signing. Some person does their signature in an elaborated manner, and some do in a short way. Based on this intuition, we have calculated the number of active signature pixels from each signature images. Active pixels in the signature images is given by equation \ref{eq:black}
    
     \begin{equation}
    AP=\sum_{i=1}^m\sum_{j=1}^n f(i,j)==0
    \label{eq:black}
\end{equation}
Where AP is the active pixels, $f(i,j)$ is the signature block or image. (m,n) are the size of the signature block or image.

    \item \textbf{Number of Connected Components}:  A connected component is an algorithm that is present in graph theory, where the different connected components in the image are labeled with a unique label. The number of components can also be considered as the distinguishing feature of the signature image. Some person has the habit of doing their signature in pieces. To calculate connected components, we have utilized 8-connectivity among the pixels in the signature image. The algorithm to extract connected components is shown in algorithm \ref{algo_conn}.
\end{enumerate}
\subsubsection{local Features}\label{lfeatures}
For each block, along with the number of active pixels and connected components some other features (effective mass, distance between effective mass and center of mass, isolated points, length and height of the block of the signature has been extracted.
\begin{enumerate}
    \item \textbf{Distance between the effective center and center of mass}:  Center of mass and effective center distinguished in the signature images. Based on the content and density of the signature, the effective center of the image is shifted. The effective mass dependent on the user and can be considered as the discriminative features of the signature image. We have calculated the center of mass of each block by equation \ref{mass}.
    \begin{equation}
    \begin{split}
        \forall x C_X^{mass}=\frac{_{p\in f(x,y)} x_p}{2},\\ \forall y C_Y^{mass}=\frac{_{p\in f(x,y)} y_p}{2}    
        \end{split}
        \label{mass}
    \end{equation}
    Where p is the pixel value, f(x,y) is the image block and $x_p$, $y_p$ is the pixel in x and y direction respectively.
     Based on the content present in the block we have calculated the effective center. Effective center has been calculated with equation \ref{effective}.
     
     \begin{equation}
     \begin{split}
        C_X^{effec}=\frac{max(\sum x_i==0)}{2},\\
        C_Y^{effec}=\frac{max(\sum y_j==0)}{2}    
         \end{split}   
     \label{effective}
     \end{equation}
     Where i=1,2,..m and j=1,2,...n.\\
     Distance (d) between these center of mass and effective center is calculated by equation \ref{dist}
         
         \begin{equation}
            d=\sqrt{(C_X^{effec}-C_X^{mass})^2+(C_Y^{effec}-C_Y^{mass})^2}        \label{dist}
         \end{equation}


    \begin{algorithm}[!ht]
\SetAlgoLined
\KwOut{Number of Connected Component}
\KwIn{Signature Image}

 1. Select one pixel in the image, and assign a label say 1.\;
2. \eIf{The selected pixel is a foreground, and it is not labeled} { Assign it the current label and put the pixel in the queue of the first component.
       }
       {
            The pixel belongs to the background, or it is already labeled, then repeat step 2 until the next unlabeled pixel
        }
3. Take out one element from the queue and check its label based on 8-connectivity.\;
    \If{The neighbor belongs to the foreground and not labeled} {assign the pixel to the current label and insert it in the queue}
Repeat step 3 until the end of the queue.\;
4. Repeat step 2 for the different pixels in the image, and the label will increase by one each time for a different component.

 \caption{Algorithm to find Connected Components}
\label{algo_conn}
\end{algorithm}

    \item \textbf{Isolated point}: Sometimes the signer draws some points in the signature, and these points can also contribute to calculating a robust feature set of the signature image. We have considered a $3 X 3$ neighborhood to find the isolated point at each pixel location.


    \item \textbf{Length and Height of the signature}: The length and height of the signature vary from person to person. Some people sign with their initials, and some signs with their names. These pieces of information can also be considered as the feature of the signature. Based on the pixel in the x and y directions, the height and length of the signature have calculated.
    \begin{equation}
    \begin{split}
      Height =max(\sum x_i==0); \quad i=1,2,...m\\
       Length =max(\sum y_j==0); \quad j=1,2,...n
        \label{lH}
    \end{split}
    \end{equation}
    Where $\sum x_i$ and $\sum y_j$ is the sum of the active pixels in $i^{th}$ row and $j^{th}$ column respectively.
\end{enumerate}


\section{Experiment}

 We have proposed this method to know more about the trajectory of the signature images. From each signature image, the feature described in section \ref{gfeatures} and \ref{lfeatures} has been extracted. To classify these features in the correct classes, we have utilized the power of Artificial Neural Network. With the extensive experiment, we have fixed the number of hidden neurons to 200 in one hidden layer, and at the output layer, we have used Softmax as the transfer function. Softmax gives the probability of the sample belonging to each class. Among all the classes, the class has the highest probability; the sample belongs to that class. In this experiment, Neural Network uses a supervised Scaled Conjugate Gradient (SCG)\cite{scg} as a training function with hold one out-validation. SCG provides a benefit while working in low memory situations as it takes less memory for training of the network. With the advantage of the low memory requirement the proposed algorithm can be used for the real time applications. In hold one out validation, a part of the feature set has been kept separately for the purpose of validation of the training process.
 
 This experiment has been carried out on MATLAB 2019a on a personal computer with 16GB RAM and i7 processor.
 
 \subsection{Dataset}\label{dataset}
 In this work we have used different datasets. Those are described in the following section. 
\begin{enumerate}
    \item MCYT-100: This dataset contains signature of 100 individuals and each signer has 25 genuine and 25 forged images. Since this is an online dataset, it has additional information like coordinate (x, y), pressure, azimuthal angle etc. For this experiment we have considered only the image of the signature not the additional information \cite{mcyt1}.
    \item MCYT-75: This is the offline version of the MCYT-100 dataset \cite{mcyt1,mcyt2}. This has 75 signer and each signer has 15 genuine and 15 forged images. This dataset do not have any auxiliary information as MCYT-100 has because it has been collected in the offline mode.
    \item GPDS: This is the largest available dataset till date \cite{GPDS1,GPDS2}. It has 4000 individuals and each has 24 genuine and 30 forged images. For this experiment we have considered only 300 signers.
    
    \item BHSig260: This dataset has the collection of the signatures in two different regional languages one is Bengali and other is Hindi \cite{pal}. In Bengali there were 100 signer were involved and in Hindi dataset 160 was involved. Each signer has 24 genuine and 30 forged images in both the regional language dataset.
    
    \item CVBLSig: This dataset is collected through Computer Vision Biometric Lab in two sessions. The first session has 137 individuals, in the second session, 467 persons were involved. In the first session, each signer has 20 images, and in the second session, each person has 15 images. There were no forged signatures collected in this dataset.
\end{enumerate}
This experiment has been carried out on skilled forgery along with random forgery. The proposed method can distinguish between genuine and skilled forged samples without being trained with skilled forged samples.



\section{Results}
This work is focused on designing a signature verification method that uses geometrical features of the signature images. 

\begin{figure*}[!ht]
     \includegraphics[width=\linewidth]{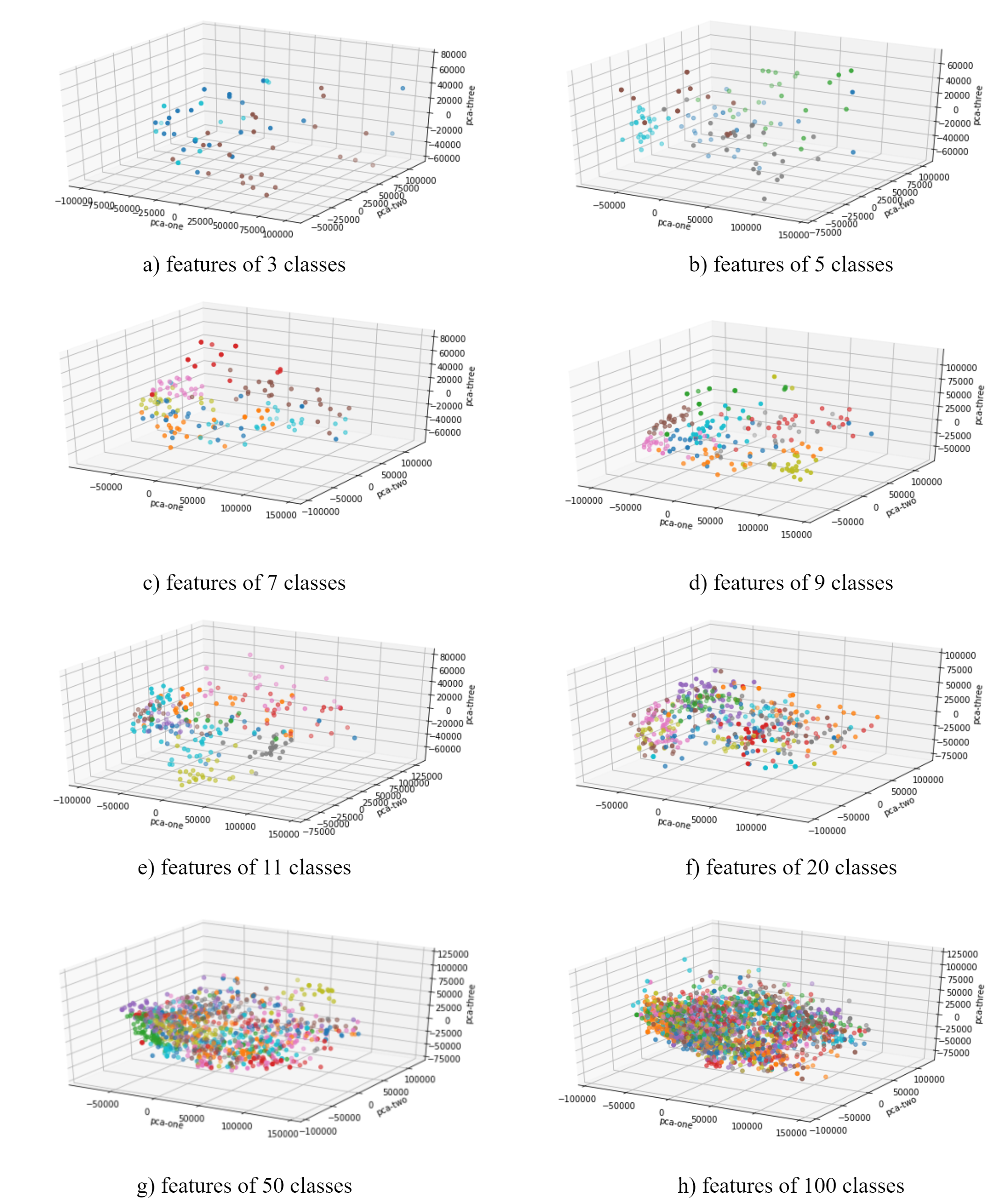}
    \caption{Visualization of the features of Bengali dataset}
    \label{fig:featureVisual}
\end{figure*}

Figure \ref{fig:featureVisual} shows the visualization of the features on Bengali dataset on different number of signers. 
The plots are shown after applying the PCA. From the dimension of the 202, three principal components are selected, and the variation between the features has been shown in the figure \ref{fig:featureVisual}.
It is visible from the figure \ref{fig:featureVisual} that the features are not linearly separable, so the distance based or linear kernel will not work for this problem, that is the reason that we have used ANN in this experiment.

\par Features discussed in section \ref{gfeatures} and \ref{lfeatures} has been extracted from both genuine and skilled forged images. With the genuine image features, the neural network has been trained and validated, and for testing, we have used features of skilled forged images. It is clear from Table \ref{table:acc_proposed} on the Hindi dataset the proposed method gives better performance for random as well as skilled forged images.

 \begin{table}[H]

\begin{center}
\begin{tabular}{|l|c|c|c|}
\hline
Dataset   & $Accuracy_{RF}$ in (\%) & $Accuracy_{SF}$ in (\%)\\
\hline
MCYT-100  &  97.36 & 79.32\\
MCYT-75  & 97.33 & - \\
GPDS300  & 92.32 & 83.2 \\
BHSig Bengali& 97.79 & 76.03\\
BHSig Hindi  & 95.29 & 83.5\\
CVBLSig-V1 & 97.55 & -\\
CVBLSig-V2 & 83.38 & -\\

\hline

\end{tabular}
\end{center}

\caption{Accuracy with the Proposed method}
\label{table:acc_proposed}
\end{table}

Table \ref{table:acc_proposed} shows the performance of the proposed method on different datasets. We have received 97.36\% and 97.33\% accuracy on MCYT-100 and MCYT-75 dataset, respectively. This experiment has also been carried out on two regional language signatures Bengali and Hindi. The proposed method gives 97.79\% and 95.29\% accuracy on the Bengali and Hindi dataset, respectively. Artificial Neural Network has been trained three times with the feature set extracted through the feature extraction phase, and Table \ref{table:acc_proposed} shows the average accuracies. This experiment has also been carried out on skilled forged signatures, and performance of the same is described in Table \ref{table:acc_proposed}. It is clear from the Table \ref{table:acc_proposed} that BHSig Hindi dataset outperformed the proposed method when skilled forgery has been included because the genuine and forged images has the more distinguished features.
We have also experimented on the locally collected dataset CVBLSig-V1 and CVBLSig-V2 and received accuracy of 97.55\% and 83.38\%, respectively. The performance of the proposed method on CVBLSig-V2 is lower, because there is the inter-class similarity between the features in the dataset. Since this dataset does not contain any forged samples, so it is not possible for checking against the skilled forgery.

In literature Equal Error Rate (EER) has been considered as the comparative measure for MCYT-100 and MCYT-75, and accuracy for the BHSig260 dataset. To show the fair comparison, we have calculated accuracy and EER for BHSig260 and MCYT dataset, respectively. 

\subsection{Comparison and Discussion}
This section describes the comparison between the proposed method and state of the art methods that were introduced in the literature. Some methods follow the function-based approach, which includes Dynamic Time Warping (DTW), Gaussian Mixture Model (GMM), etc., and some used the descriptor-based method. The proposed method is quite simple and effective. In this paper, geometrical features have been utilized to provide authentication to the signer. 
\par Table \ref{table:Performanceexisting} shows the comparative analysis on the dataset MCYT-100 and MCYT-75. It is clear from the Table \ref{table:Performanceexisting} that our method outperformed as compared to \cite{cf+tf,10,12}. Manjunath et al. has reported EER 0.80\% on MCYT-100 dataset \cite{10}. They have used a writer dependent method; for each writer, they have trained a different classifier. By doing this, they have increased the complexity of the method. 
In this experiment, we have utilized the geometrical information present in the signature images for verification and received good performance as compared to most of the state of the art methods.

 
 \begin{table}[!ht]
\centering


{\begin{tabular}{|p{1.5cm}|p{4.5cm}|p{1cm}|}
\hline
Database & Model  & EER \\
\hline
\multirow{5}{*}{MCYT-100} & W-Dependent feature \cite{10} &  7.75 \\
 &  W-Dependent feature (NN+PCA) \cite{10}&   0.80\\
 & DTW and GMM \cite{12}& 2.12\\
  & CF+TF \cite{cf+tf} & 1.20\\
& \textbf{Proposed}  & \textbf{0.30} \\
\hline
\multirow{8}{*}{MCYT-75}&{GLCM}\cite{14} & {2.30} \\
& {GLCM+WT}\cite{14}  &{2.44} \\
&{FV with fused KAZE features}\cite{OKAWA}& 5.47\\
&{Fixed size representation}\cite{haff}&{3.64}\\
&{Deep Multimetric learning}\cite{amir}&{1.73}\\
&{VLAD and KAZE}\cite{OKAWA}&{6.4}\\
&{WI using asymmetric pixel relation}\cite{121}&3.5\\
&\textbf{Proposed}  &\textbf{0.31} \\
\hline
\end{tabular}}{}
\caption{EER Comparison on MCYT Dataset\label{table:Performanceexisting}}
\end{table}

 Table \ref{table:Performanceexisting} also shows the comparative analysis of the MCYT-75 dataset. Our method outperformed many state of art methods reported in \cite{14,amir,OKAWA,haff,121}. Table \ref{table:Performanceexisting} shows that MCYT-100 has the lower EER than the MCYT-75 dataset on proposed method, the possible reason for that is the accuracy received using MCYT-100 dataset is more than the MCYT-75 dataset. We have achieved the higher accuracy on MCYT-100 dataset because there is no noise introduced in the acquisition process of the MCYT-100 dataset. Table \ref{table:PerformanceexistingBHSig} shows the comparison on the dataset BHSig260. It is visible from the table \ref{table:PerformanceexistingBHSig} that our method gives better performance as compared to the state of the art methods \cite{pal,Dutta}. The possible reason for such performance is because of the less intra-class variation and large inter-class variation in the feature set.




\begin{table}[!ht]
\centering

{\begin{tabular}{|p{2cm}|p{4cm}|p{1.2cm}|}
\hline
Database & Method  &Accuracy\% \\
\hline
\multirow{4}{*}{BHSig Bengali} &  LBP and ULBP\cite{pal} & 66.18
 \\
 &  Compact Corelated Features\cite{Dutta}&   84.90\\
 &  SigNet(Siamese)\cite{13}& 86.11\\
 
& \textbf{Proposed}  & \textbf{97.79} \\
& {Proposed(Skilled Forged)} & {76.03}\\
\hline
\multirow{3}{*}{BHSig Hindi}& LBP and ULBP\cite{pal}& {75.53} \\
&Compact Corelated Features\cite{Dutta} & {85.90} \\
& SigNet(Siamese)\cite{13}  &{84.64} \\
&\textbf{Proposed}  &\textbf{95.29} \\
&{Proposed(Skilled Forged)}  &{83.5} \\
\hline
\end{tabular}}{}
\caption{Accuracy Comparison on the BHSig260 Dataset\label{table:PerformanceexistingBHSig}}
\end{table}

         

\begin{figure}[!ht]
    \centering
    \includegraphics[width=\linewidth]{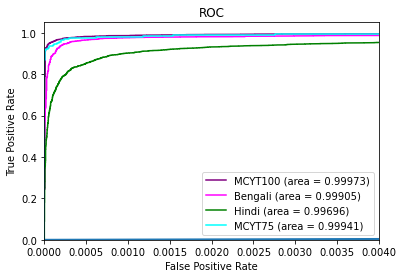}
    \caption{ROC Curve for the different datasets}
    \label{fig:roc}
\end{figure}

Figure \ref{fig:roc} shows the ROC (Receiver operator characteristics) curve on different datasets (MCYT-100, MCYT-75, Bengali, and Hindi). ROC shows the power of discrimination among the samples in the dataset. It has been noticed from the figure \ref{fig:roc} that the area under the curve (AUC) is maximum for the dataset MCYT-100. Because of the higher AUC the MCYT-100 dataset is achieving lower EER presented in the Table \ref{table:Performanceexisting}. 

\section{Conclusion and Future work}
In the field of computer vision, signature verification considered a very challenging problem because of its intra-class variation and inter-class similarity. Through this paper our objective is to provide a signature verification methods that can be used in real time and perform better. In this paper we have proposed a signature verification method that is based on the geometrical features. with the geometrical features from the signature images, we have drawn a robust feature set based on some simple characteristics from the signature images. With the proposed method we are successful in reducing the EER on MCYT-100 to 0.30\% and on MCYT-75 to 0.31\%. The accuracy recorded on the dataset BHSig Bengali and Hindi is 97.79\% and 95.29\%, respectively. On locally collected dataset CVBLSig-V1 and CVBLSig-V2, we have achieved 97.55\% and 83.38\%, respectively.

This method works on a pixel level; it requires a long time to extract the features from the signature image. On the other hand, for training the Neural Network, the required time is less.

\section*{Funding}
Not Applicable.
\section*{Conflict of Interest}
The authors declare that they have no conflict of
interest.
\section*{Availability of data and material}
Once the manuscript is accepted, we will release the data used in the manuscript.

\section*{Code availability}
Once the manuscript is accepted, we will release the code used in the manuscript.

\bibliographystyle{spmpsci}
\bibliography{ref}

\end{document}